\documentclass[sigconf]{acmart}
\AtBeginDocument{%
  }

\acmYear{2025}\copyrightyear{2025}
\setcopyright{acmlicensed}
\acmConference[SoCC '25]{ACM Symposium on Cloud Computing}{November 19--21, 2025}{Online, USA}
\acmBooktitle{ACM Symposium on Cloud Computing (SoCC '25), November 19--21, 2025, Online, USA}
\acmDOI{10.1145/3772052.3772239}
\acmISBN{979-8-4007-2276-9/25/11}

\usepackage{algorithm}
\usepackage{algorithmic}
\usepackage{enumitem}
\usepackage{graphicx}
\usepackage{makecell}
\usepackage{xspace}
\usepackage{hyperref}
\usepackage{balance}
\acmSubmissionID{179}

\AtEndPreamble{%
  \theoremstyle{acmdefinition}

}

\begin{document}

\newcommand{\SysName}{\text{AdaSpec}\xspace}
\newcommand{\SysNameMinus}{\text{AdaSpec-}\xspace}
\title{\SysName: Adaptive Speculative Decoding for Fast, SLO-Aware Large Language Model Serving}

\author{
    Kaiyu Huang\textsuperscript{1,2},
    Hao Wu\textsuperscript{3},
    Zhubo Shi\textsuperscript{1},
    Han Zou\textsuperscript{1},
    Minchen Yu\textsuperscript{2,4*},
    Qingjiang Shi\textsuperscript{1,2*}
}
\affiliation{%
  \institution{
    \textsuperscript{1}Tongji University
    \country{China}
  }
  \institution{
    \textsuperscript{2}Shenzhen Research Institute of Big Data, The Chinese University of Hong Kong, Shenzhen
    \country{China}
  }
  \institution{
    \textsuperscript{3}Huazhong University of Science and Technology
    \country{China}
  }
  \institution{
    \textsuperscript{4}School of Data Science, The Chinese University of Hong Kong, Shenzhen
    \country{China}
  }
}
\thanks{*Corresponding authors. E-mail addresses: yuminchen@cuhk.edu.cn (Minchen Yu), shiqj@tongji.edu.cn (Qingjiang Shi)..}


\begin{abstract}
Cloud-based Large Language Model (LLM) services often face challenges in achieving low inference latency and meeting Service Level Objectives (SLOs) under dynamic request patterns. 
Speculative decoding, which exploits lightweight models for drafting and LLMs for verification, has emerged as a compelling technique to accelerate LLM inference.
However, existing speculative decoding solutions often fail to adapt to fluctuating workloads and dynamic system environments, resulting in impaired performance and SLO violations. 
In this paper, we introduce \SysName, an efficient LLM inference system that dynamically adjusts speculative strategies according to real-time request loads and system configurations. 
\SysName proposes a theoretical model to analyze and predict the efficiency of speculative strategies across diverse scenarios. 
Additionally, it implements intelligent drafting and verification algorithms to maximize performance while ensuring high SLO attainment. 
Experimental results on real-world LLM service traces demonstrate that \SysName consistently meets SLOs and achieves substantial performance improvements, delivering up to 66\% speedup compared to state-of-the-art speculative inference systems. The source code is publicly available at \url{https://github.com/cerebellumking/AdaSpec}
\end{abstract}






\maketitle

\section{Introduction}

Large Language Models (LLMs) have demonstrated remarkable versatility across a broad spectrum of tasks~\cite{NEURIPS2020_1457c0d6}, such as translation~\cite{lu-etal-2024-llamax}, question answering~\cite{10.1162/tacl_a_00276}, and mathematical reasoning~\cite{cobbe2021training}. 
This success has fueled the rapid and widespread adoption of cloud-based LLM services, which enable users to submit real-time queries and receive generated responses via LLM inference. 
The inference process of LLMs typically consists of two main phases: the prefill phase, where the model processes the user’s input prompt and initializes the context, and the decoding phase, where the LLM generates output tokens sequentially. The decoding phase generally dominates the end-to-end request latency, particularly for long output sequences.
Therefore, a critical requirement for LLM services is to guarantee fast and consistent token generation performance, which is often specified as Service-Level Objectives (SLOs) such as a maximum allowable Time Per Output Token (TPOT)~\cite{10.1145/3698038.3698523,10.1145/3472883.3486993}.

Speculative decoding has recently emerged as a compelling solution for accelerating LLM inference~\cite{pmlr-v202-leviathan23a,chen2023accelerating}.  
This approach leverages a lightweight draft model to produce multiple candidate tokens, which are then collectively verified by the target LLM. 
By generating and validating multiple tokens at once, speculative decoding produces the same outputs as traditional autoregressive decoding, but greatly reduces the overhead associated with invoking the LLM for each token. This efficiency leads to improved GPU utilization, lower inference latency, and higher overall system throughput. 

However, existing speculative decoding solutions often face notable challenges in real-world cloud scenarios, where request patterns and resource configurations are highly dynamic. 
Most current solutions rely on a pre-defined speculative length---the number of draft tokens generated per speculative round. 
While increasing this speculative length can raise the likelihood of verifying more tokens in a single pass, it also amplifies the risk of token rejections, leading to wasted computation and higher end-to-end latency~\cite{liu2024optimizing}. 
Finding the optimal speculative length is complicated, as it depends on a complex interplay of factors such as request rates, model loads, and hardware resources (see \S\ref{sec:limitations}).
Consequently, static approaches that use a fixed speculative length fails to consistently deliver satisfactory performance under changing workloads and environments.

Moreover, existing solutions often overlook token-level SLOs in LLM inference, leading to inconsistent SLO attainment. Although speculative decoding reduces inference latencies compared to the autoregressive approach, the variability in token acceptance during the verification process can cause significant fluctuations in overall token generation latency, such as TPOT. This inherent variability poses challenges for speculative decoding to consistently satisfy SLO requirements, undermining the ability to deliver reliable and predictable quality of service to users.

To address these limitations, an efficient speculative inference system should enable fine-grained, adaptive control over speculative lengths, remaining responsive to dynamic workloads and SLO requirements. 
Following this principle, we present \SysName, a system that leverages adaptive speculative decoding to improve performance and SLO attainment for LLM inference. 
\SysName aims to dynamically configure the optimal speculative length for each request, adapting to fluctuating request patterns, diverse system configurations, and varying SLO constraints.
However, achieving adaptive speculative decoding in \SysName presents three challenges. 

The first challenge is to accurately characterize the efficiency of speculative decoding under real-world, dynamic conditions. 
In cloud environments, LLM inference workloads can vary significantly in batch size, context length, and workload heterogeneity. 
While system-level characteristics such as batch size and context length are easy to monitor, the acceptance rate of draft tokens is difficult to assess during drafting. 
This challenge stems from the inherent uncertainty: we can only detect whether drafted tokens will be accepted until the verification phase.
Without a reliable efficiency estimation mechanism, selecting the optimal speculative length becomes infeasible.
To address this, \SysName introduces an efficiency model (see \S\ref{sec:efficiency_model}) that estimates both the expected acceptance rate and execution time, leveraging real-time draft token confidence scores and current system configurations. This model empowers the adaptive drafter (see \S\ref{sec:adaptive_drafter}) to dynamically adjust the speculative length at the batch level. Through this approach, \SysName achieves more informed and responsive control over speculative decoding, significantly enhancing overall system performance.

Second, enabling fine-grained and flexible control of speculative length at the request level presents significant challenges. 
In practice, LLM inference systems batch heterogeneous user requests to maximize throughput and resource utilization. However, existing speculative decoding methods generally apply a uniform speculative length for all requests within a batch, overlooking substantial differences in token acceptance probabilities among queries. 
This rigid approach can result in inefficiencies---for example, applying the same speculative length for both simple and complex prompts may cause unnecessary token rejections or insufficient acceleration for less demanding requests.
To tackle this challenge, \SysName introduces the confidence prior verifier (see \S\ref{sec:confidence_prior_verifier}), which decouples the drafting and verification stages of decoding. 
By leveraging per-request acceptance probability estimates, the system can selectively allocate verification efforts to tokens with higher predicted acceptance rates. This enables each request in a batch to effectively utilize its own optimal speculative length, without requiring changes to the overall batch execution pipeline.

Third, \SysName must balance the efficiency gains of speculative decoding with the need to meet SLOs. Speculative decoding introduces probabilistic performance variations: while longer speculative lengths can boost throughput, they also increase the likelihood of token rejections and latency spikes, which may result in SLO violations. Existing approaches often neglect this crucial trade-off, leading to degraded service quality under strict latency constraints.
\SysName addresses this by incorporating a SLO-aware efficiency estimator (see \S\ref{sec:efficiency_estimator}) that proactively evaluates whether a chosen speculative decoding configuration can satisfy TPOT requirements before execution. This enables \SysName to make adaptive, real-time decisions that optimize for both high performance and SLO compliance, ensuring robust efficiency without sacrificing reliability.


We have implemented \SysName atop vLLM~\cite{10.1145/3600006.3613165} and evaluated using real-world LLM inference traces.
Evaluation shows that \SysName can adapt to varying workloads and system configurations, consistently achieving low latency and high SLO attainments.
Compared with state-of-the-art speculative inference systems, \SysName enables up to 66\% speedups while meeting SLO requirements.
  
To summarize, we make the following contributions:
\setlist[itemize]{itemsep=0ex,topsep=0ex,parsep=0ex}
\begin{itemize}
\item We proposed \SysName, an efficient LLM inference system with adaptive speculative decoding.
\item We designed a fine-grained control mechanism for speculative length (i.e., adaptive drafter and confidence prior verifier), which can adapt to varying request patterns and system configurations and achieve optimal performance with theoretical guarantee.
\item We designed an efficient algorithm  (i.e., SLO-aware efficiency estimator) to achieve high SLO attainment under varying performance requirements. 
\item We evaluated \SysName using real-world production traces, which demonstrates that \SysName can improve inference performance by up to 66\% compared with baselines.
\end{itemize}

\section{Background and Motivation}

\subsection{LLM Inference Service}

\begin{figure}[t]
\centering
\includegraphics[width=\linewidth]{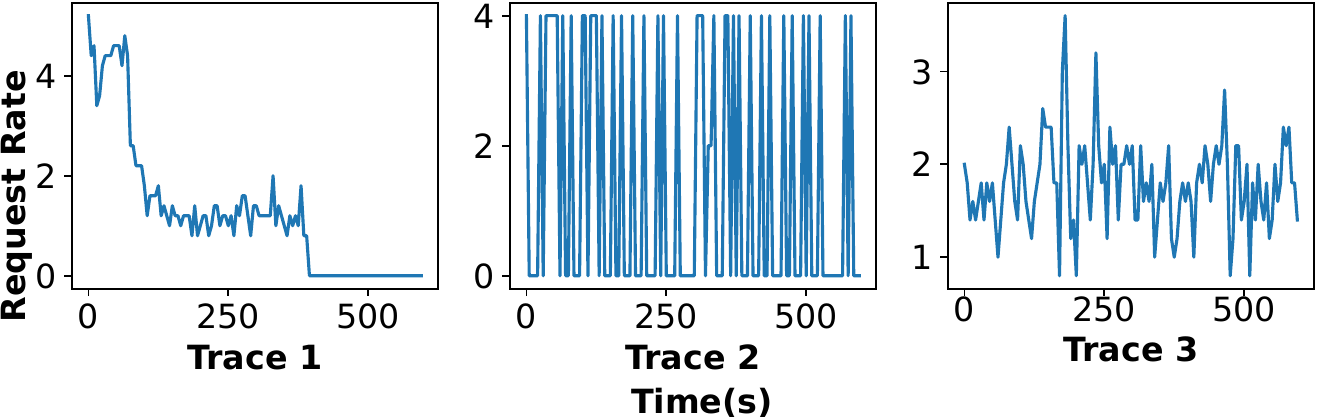}
\caption{Different request patterns in real production traces.}
\label{fig:12}
\end{figure}

In LLM inference~\cite{achiam2023gpt,bengio2000neural}, the model accepts a list of tokens $\mathbf{x}=(x_1,\ldots,x_n)$, and outputs the conditional probability distribution of the next token $(x_{n+1} \vert x_1,\ldots,x_n)$. 
Trained LLMs are deployed onto inference systems to provide online services, where they receive requests consisting of token lists and generate a complete output sequence $(x_{n+1},\ldots,x_{n+T})$ by predicting the next token $T$ times.

In real-world LLM inference systems, they usually need to handle dynamic request arrivals from a variety of workloads. 
Fig. \ref{fig:12} displays three representative LLM inference traces from BurstGPT~\cite{burstgpt} and Mooncake~\cite{qin2025mooncake}. 
The first pattern (trace 1) exhibits substantial request rate variations, demonstrating significant minute-level fluctuations in system load. 
The second pattern (trace 2) reveals high-frequency periodic oscillations, indicating batch-style request arrivals that form intensive yet regular load patterns. 
The third pattern (trace 3) shows rapid fluctuations around a central value, with second-level dynamics that reflect both the stochastic nature and instantaneous variability of workload.
Collectively, these distinct patterns capture the diversity and complex dynamics of real-world production system loads.
A key requirement of LLM inference is to meet stringent latency requirements (i.e., SLOs), which necessitates inference systems to efficiently handle dynamic requests and achieve low-latency token generation~\cite{fu_serverlessllm_2024,234998,zhang2023shepherd,faaswap}.

Request batching is a common approach to improve inference performance and resource efficiency for LLM inference~\cite{yu2022orca,gao2018low}.
It fuses multiple requests associated with the same model as a batch, and executes them together to fully exploit parallel computing resources of hardware accelerators such as GPUs. 
Mainstream LLM inference systems such as TensorRT-LLM~\cite{TensorRT-LLM} and vLLM~\cite{10.1145/3600006.3613165} offer the ability of batching for enhanced efficiency.
While request batching enhances GPU utilization during decoding, this method fails to address inefficiencies in scenarios with persistently low per-model request rates. 
According to real-world inference traces (Fig.~\ref{fig:1}), LLM services frequently experience prolonged periods of low request rates---a pattern consistent with prior observations of inference workloads\cite{faaswap}.
These findings underscore a critical requirement for LLM inference systems: maintaining high computational efficiency even under sustained low request arrival rates.

\subsection{Speculative Decoding}

\begin{figure}[t]
\centering
\includegraphics[width=\linewidth]{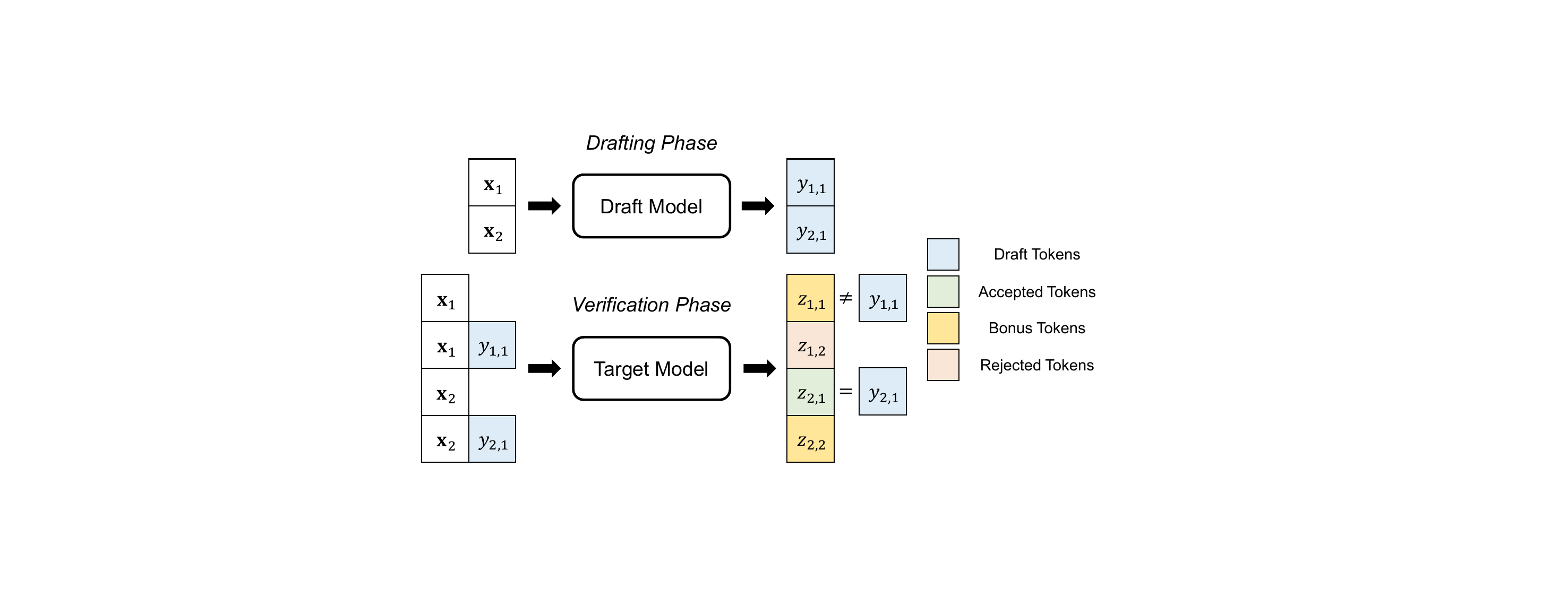}
\caption{A single-step speculative decoding process with a speculative length of 1. We primarily discuss the greedy sampling, where the token with the highest confidence score is selected as the output instead of the whole distribution.}
\label{fig:1}
\end{figure}

\noindent Speculative decoding addresses the inefficiency of autoregressive decoding in LLM inference systems.
This approach utilizes a small draft model to predict the following tokens and verify them at once, improving resource efficiency and end-to-end performance. As shown in Fig. \ref{fig:1}, speculative decoding has two parts: drafting and verification. During the drafting phase, given a list of tokens $\mathbf{x}=(x_1,\ldots,x_n)$, speculative decoding uses an efficient draft model to generate a draft sequence $\mathbf{y}=(y_1,\ldots,y_{SL})$, where $SL$ is the speculative length. In the verification process, the draft sequence is input into the target model in parallel for a single forward pass, generating multiple output distributions $P(z_1 \vert x_{1:n}),\ldots,P(z_{SL+1} \vert x_{1:n},y_{1:SL})$. Each drafted token $y_t$ is compared with $z_t$: if $y_t = z_t$, the token is accepted; otherwise, verification stops at position $m = t$, where $z_m \ne y_m$ is the first rejected token. The final output consists of the accepted draft tokens $(y_1, \ldots, y_{m-1})$ and one bonus token $z_m$ directly from the target model, forming the output $(y_1, \ldots, y_{m-1}, z_m)$. Since all accepted tokens and bonus tokens come from the target model’s predictions, speculative decoding guarantees lossless equivalence to standard autoregressive decoding.

The speculative length, by controlling the number of candidate tokens generated by the draft model, directly impacts both the latency benefits and resource consumption of speculative decoding. 
Mainstream methods~\cite{pmlr-v202-leviathan23a,NEURIPS2023_6034a661,10.1145/3620666.3651335,pmlr-v235-cai24b,lieagle} rely on offline analysis to determine a static speculative length, with its optimal value typically identified through heuristic search on specific datasets and under fixed hardware configurations.
Furthermore, while a few methods, such as SmartSpec~\cite{liu2024optimizing}, are capable of setting the speculative length for each decoding step, their coarse-grained control still fails to adapt to the fine-grained variations in real-world request scenarios.

\subsection{Limitations of Existing Solutions}
\label{sec:limitations}

Current speculative decoding approaches face three limitations in the practical deployment of LLM inference systems in the cloud.
These limitations stem from their lack of adaptability to dynamic request patterns, system configurations, and SLO constraints.

\begin{figure}[t] 
\centering
\includegraphics[width=\linewidth]{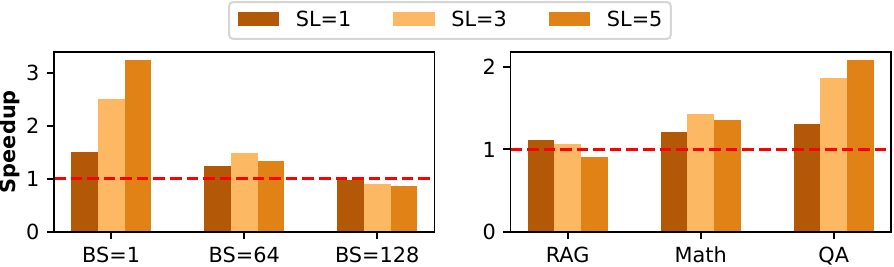}
\caption{Speedup across various speculative lengths under diverse request quantities and types. The red dashed line represents the speedup value of 1.}
\label{fig:2}
\end{figure}

\noindent\textbf{Limitation \#1: Inefficiency under dynamic workloads.} 
We tested the speedup ratio of speculative decoding compared to autoregressive decoding under different request quantities (batch sizes of 1, 64, 128) and different request types (retrieval-augmented generation, mathematical reasoning, question answering), with identical system configurations across various speculative lengths.
As shown in Fig. \ref{fig:2}, conventional speculative decoding methods, which rely on fixed speculative lengths, fail to adapt to dynamic changes in request patterns.
Under high-load scenarios (batch size 128), excessively long speculative lengths lead to: (1) increased computational overhead during the verification phase due to redundant tokens generated by the draft model; (2) resource contention caused by batch sizes exceeding device memory.
Conversely, under low-load scenarios (batch size 1), excessively short speculative lengths fail to fully utilize idle computational resources, leaving the acceleration potential of speculative decoding untapped.
Workload characteristics represent another critical variable. We demonstrate this by analyzing workloads extracted from RAG~\cite{karpukhin-etal-2020-dense}, mathematical reasoning~\cite{cobbe2021training}, and question-answering~\cite{10.1162/tacl_a_00276} datasets, each exhibiting different optimal speculative lengths.
Evidently, applying the same speculative length across different requests results in both redundant token generation and wasted acceleration potential, making it difficult to achieve optimal latency optimization.

\begin{figure}[t]
\centering
\includegraphics[width=\linewidth]{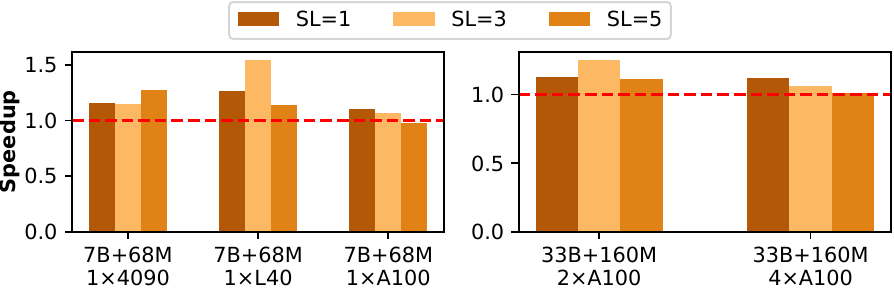}
\caption{Speedup across different speculative lengths for various draft-target model pairs on diverse computing platforms.}
\label{fig:3}
\end{figure}

\noindent\textbf{Limitation \#2: Agnostic to system configurations.}
We evaluated the speedup ratio of speculative decoding compared to autoregressive decoding across different draft-target model pairs deployed on varying hardware configurations (see Tab. \ref{tab:1}) under multiple speculative lengths.
As shown in Fig. \ref{fig:3}, when hardware configurations (e.g., GPU models and counts in the computing platform) or model pairings (combinations of draft and target models) vary, the fixed-length strategy employed in traditional speculative decoding loses its optimality. 
The root cause lies in shifts in the hardware roofline model: when the target model’s arithmetic intensity falls below the platform’s balance point, the parallel token processing in the verification phase can significantly improve computational efficiency~\cite{marzollo2024sssd}. 
However, under compute-bound conditions, the probabilistic nature of speculative decoding (where proposed draft tokens may be rejected) reduces its token generation efficiency compared to autoregressive decoding, leading to latency degradation. 
This configuration sensitivity highlights the method’s inability to dynamically adapt to changing system environments.

\begin{figure}[t]
\centering
\includegraphics[width=\linewidth]{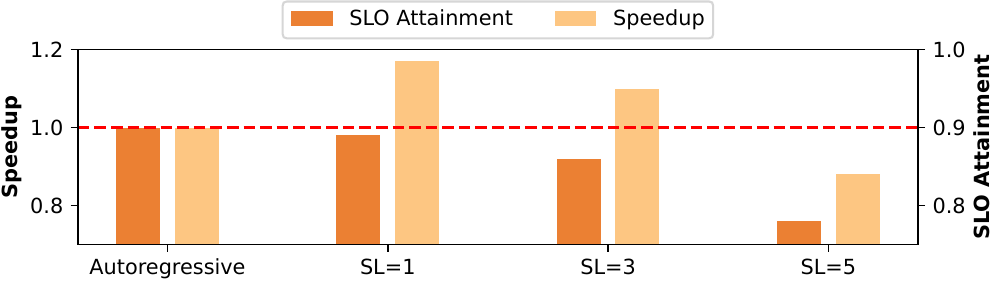}
\caption{Relationship between SLO attainment and end-to-end latency speedup under different speculative lengths. The red dashed line represents the speedup value of 1 and the SLO attainment value of 0.9.}
\label{fig:4}
\end{figure}

\noindent\textbf{Limitation \#3: SLO violations.} 
We evaluated the SLO attainment of speculative decoding under different speculative lengths, using the P90 TPOT of autoregressive decoding as the threshold.
As revealed in Fig. \ref{fig:4}, while speculative decoding can achieve higher acceleration ratios compared to autoregressive decoding, its SLO attainment may be significantly lower. 
This issue arises from the probabilistic variability of speculative decoding's acceleration effects: longer speculative lengths increase the expected acceleration ratio but also raise the risk of exceeding the TPOT limit (e.g., when the number of tokens actually generated is lower than expected, the TPOT for that speculative decoding attempt will significantly exceed its expected value). 
This uncertainty prevents speculative decoding from simultaneously meeting the high-performance and high-reliability requirements of inference systems.

\subsection{Key Insight and Challenges}

\noindent\textbf{Key Insight: Fine-grained dynamic control.} 
These limitations discussed in \S\ref{sec:limitations} reveal a core contradiction: the potential benefits of speculative decoding are constrained by its rigid execution framework.
We propose a solution which transforms the speculative length from a preset parameter into a dynamically controlled variable.
Specifically, the system needs to dynamically adjust the speculative length for each decoding step in real-time, based on workload characteristics, system configurations, and SLO constraints. 
This resource scheduling mechanism enables speculative decoding to both mitigate the risk of resource overload (e.g., by shortening speculative lengths under high-load conditions) and maximize computational resource utilization (e.g., by extending speculative lengths under low-load conditions), thereby overcoming the performance boundaries of static strategies.
However, achieving fine-grained dynamic control of speculative length faces three major challenges.

\noindent\textbf{Challenge \#1: How to accurately characterize the dynamic relationship between efficiency and speculative length?}
The relationship between the efficiency of speculative decoding and speculative length is influenced by dynamic request patterns and diverse system configurations in real-world scenarios (\S\ref{sec:limitations}). 
To accurately characterize this relationship, we need to construct an effective and general-purpose efficiency model for speculative decoding.
This model can guide the design of speculative strategies under varying scenarios for improved overall efficiency.

\noindent\textbf{Challenge \#2: How to timely and flexibly control the speculative length?}
The root of this challenge lies in two factors: (1) Lag in efficiency estimation: The estimation of decoding efficiency depends on the execution results of the draft model. 
This dependency prevents the system from preemptively evaluating the validity of a given speculative length before generating the draft, leading to delay in controlling. 
(2) Rigid system architecture constraints: All requests within the same batch have the same number of draft model executions, which results in the speculative length being the same for all requests, thereby neglecting fine-grained control.

\noindent\textbf{Challenge \#3: How to balance the efficiency of speculative decoding and SLO attainment?}
Due to probabilistic fluctuations in the acceleration effects of speculative decoding, there is a conflict between maximizing the efficiency of speculative decoding and meeting SLO requirements (\S\ref{sec:limitations}).
Therefore, we need to incorporate SLO constraints and the risk of TPOT exceeding expectations into the efficiency model, to balance acceleration gains and SLO violation risks in real-time when controlling speculative length.

\begin{figure*}[t]
\centering
\includegraphics[width=\textwidth]{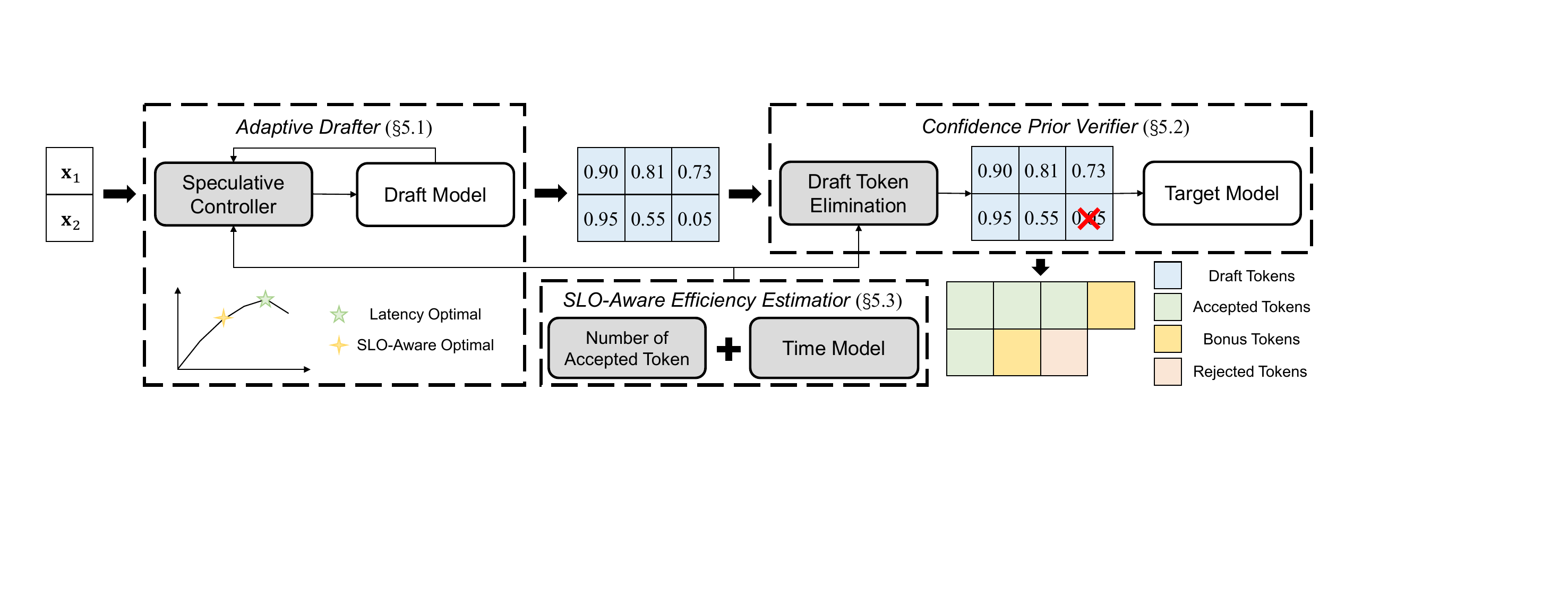}
\caption{An overview of \SysName's modules: adaptive drafter, confidence prior verifier, and SLO-aware efficiency estimator.}
\label{fig:5}
\end{figure*}

\section{\SysName's Overview}

\noindent We proposed \SysName, an efficient LLM inference system with adaptive speculative decoding.
To accurately characterize the dynamic relationship between efficiency and speculative length (\textbf{Challenge \#1}), we establish an effective and general-purpose efficiency model for speculative decoding (\S\ref{sec:efficiency_model}). 
Building upon this foundation, \SysName adopts a modular design comprising three key components: adaptive drafter, confidence prior verifier, and SLO-aware efficiency estimator. 
The design principles of these components and their collaborative operational mechanism are illustrated in Fig. \ref{fig:5}.

\noindent\textbf{Adaptive Drafter (\S\ref{sec:adaptive_drafter}).} 
To achieve timely control over speculative length (\textbf{Challenge \#2-1}), \SysName introduces an adaptive drafter, integrating efficiency estimation into the drafting phase to achieve real-time, step-level speculative length control. 
As illustrated in Fig. \ref{fig:5}, the adaptive drafter consists of a \textit{speculative controller} and a \textit{draft model} governed by it. 
Before each execution of the draft model, the speculative controller uses historical confidence scores to estimate the potential efficiency gain of the current speculative length in real-time. 
The draft model executes only if a positive gain is predicted; otherwise, the execution terminates timely.
After executing the draft model, the generated confidence scores are recorded as historical data and used to calibrate the actual efficiency gain of the current length, ensuring precise control over speculative length.

\noindent\textbf{Confidence Prior Verifier (\S\ref{sec:confidence_prior_verifier}).} 
To enable flexible control over speculative length (\textbf{Challenge \#2-2}), \SysName employs a confidence prior verifier to prioritize the verification of tokens with high acceptance rates. 
This approach decouples the dependency between the drafting and verification phases, breaking the rigid system architecture constraints and achieving fine-grained, request-level speculative length control. 
As shown in Fig. \ref{fig:5}, the confidence prior verifier consists of a \textit{draft token elimination} and a \textit{target model}.
Before being input into the target model, the draft token set generated by the adaptive drafter undergoes draft token elimination, where tokens are removed in ascending order of acceptance rate until the draft token set with the highest efficiency is selected.

\noindent\textbf{SLO-Aware Efficiency Estimator (\S\ref{sec:efficiency_estimator}).} 
The key operations of \SysName (speculative controller and draft token elimination) are based on the evaluation of the efficiency of speculative decoding.
Similar to vLLM~\cite{10.1145/3600006.3613165} and Orca~\cite{yu2022orca}, we use throughput to measure decoding efficiency and propose an efficiency estimator. 
This estimator employs the efficiency model of speculative decoding to adapt to dynamic request patterns and diverse system configurations. 
To achieve SLO awareness (\textbf{Challenge \#3}), the efficiency estimator incorporates SLO constraints to support rapid trade-offs between the efficiency of speculative decoding and SLO compliance.

\section{Modeling the Efficiency of Speculative Decoding}
\label{sec:efficiency_model}

To evaluate the efficiency of speculative decoding, we adopted the same throughput metric as used in LLM inference systems~\cite{10.1145/3600006.3613165,yu2022orca}, which is defined as $Throughput = NAT/T_{sd}$. By analyzing how the number of accepted tokens ($NAT$) and the time of speculative decoding ($T_{sd}$) vary with speculative length, we show that throughput either peaks at an intermediate length or decreases monotonically, providing a theoretical basis for efficient length tuning.

\subsection{The Time for Speculative Decoding}
\label{sec:4.1}

The time for speculative decoding $T_{sd}$ includes the execution time of the draft model ($T_{d}$) and the target model ($T_{v}$), expressed as:
\begin{align}
\label{eq:t_sd}
T_{sd}=T_{d}+T_{v}
\end{align}

\noindent\textbf{Modeling the forward propagation time.} Execution time in transformer models is dominated by forward passes. 
Following ~\cite{liu2024optimizing}, we model the forward propagation cost for a batch as a linear function of the number of context tokens $N_c$ and batch tokens $N_b$.
\begin{align}
\label{eq:t_fwd}
T_{fwd}(N_c,N_b)=\alpha*N_c+\gamma* N_b+\delta
\end{align}
, where the performance coefficients $(\alpha,\gamma,\delta)$ depend on the computing platform, parallelism, and the model.

\noindent\textbf{Modeling the execution time of draft and target models.} The draft model performs $SL$ forward passes and the target model performs one forward pass over all $SL+1$ positions. Under Eq. ~\eqref{eq:t_fwd}, the execution time of draft and target models can be expressed as:
\begin{align}
\label{eq:t_d}
T_d &= \alpha_d*\sum_{i=1}^{SL} N_{c}(i) + \gamma_d*\sum_{i=1}^{SL} N_{b}(i) + SL*\delta_d \\
\label{eq:t_v}
T_v &= \alpha_v*\sum_{i=1}^{SL+1} N_{c}(i) + \gamma_v*\sum_{i=1}^{SL+1} N_{b}(i) + \delta_v
\end{align}
Here, $N_{c}(i)$ and $N_{b}(i)$ denote the number of context tokens and batch tokens in the $i^{th}$ forward pass of the draft model, and $(\alpha_d, \gamma_d, \delta_d)$ and $(\alpha_v, \gamma_v, \delta_v)$ are the performance coefficients for the draft and target models, respectively.

\noindent\textbf{Modeling the relationship with speculative length.}
To understand how speculative length $SL$ influences overall latency, we express $T_{sd}$ in terms of $SL$, assuming the average context length per request is $\overline{N_c}$ and batch size is $BS$.

\begin{lemma}
\label{lem:sd_time}
The total speculative decoding time $T_{sd}$ is a quadratic function of the speculative length $SL$:
\begin{equation}
\begin{split}
T_{sd} &= BS\big( \frac{SL^2 (\alpha_{d} + \alpha_{v})}{2}  \\
&\quad + SL(\alpha_{d}\overline{N_c} + \alpha_{v}\overline{N_c} - \frac{\alpha_{d}}{2} + \frac{\alpha_{v}}{2} + \gamma_{d} + \gamma_{v} + \frac{\delta_{d}}{BS}) \\
&\quad + \alpha_{v}\overline{N_c} + \gamma_{v} + \frac{\delta_{v}}{BS} \big)
\end{split}
\end{equation}
\end{lemma}

\begin{proof}
From Eq. ~\eqref{eq:t_sd}, the total speculative decoding time is $T_{sd} = T_d + T_v$. The expressions for $T_d$ and $T_v$ are given in Eqs. ~\eqref{eq:t_d} and ~\eqref{eq:t_v}. 

To express $T_{sd}$ in terms of speculative length $SL$ and batch size $BS$, we replace the summations $\sum_i N_c(i)$ and $\sum_i N_b(i)$ in Eqs. ~\eqref{eq:t_d} and ~\eqref{eq:t_v} using the average context length $\overline{N_c}$ and constant batch size. After simplification, we obtain the quadratic form in $SL$.
\end{proof}

This quadratic structure implies that decoding time grows super-linearly with speculative length, and thus, optimizing $SL$ is essential for balancing accuracy and latency.

\begin{figure}[t]
\centering
\includegraphics[width=\linewidth]{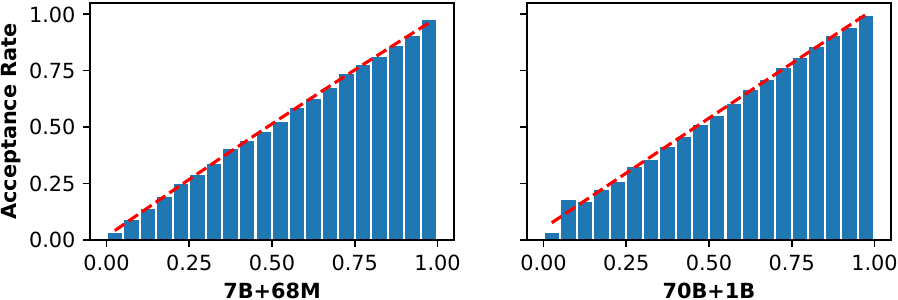}
\caption{Average acceptance rates for different confidence score intervals of the draft model. The red dashed line connects (0,0) and (1,1) to aid in visual assessment.}
\label{fig:6}
\end{figure}

\subsection{The Number of Accepted Tokens}
\label{sec:4.2}

To quantify the effectiveness of speculative decoding, we develop a model to estimate the number of accepted tokens. Due to varying system loads and request types (see \S\ref{sec:limitations}), we aim to provide a general estimation that avoids relying on request-specific conditions.

Let $BS$ denote the batch size and $SL$ the speculative length. The number of accepted tokens $NAT$ is defined as:
\begin{equation}
\label{eq:nat}
NAT = \sum_{i=1}^{BS} \left( \sum_{j=1}^{SL} AR_{i,j} + 1 \right) = \sum_{i=1}^{BS} \sum_{j=0}^{SL} AR_{i,j}
\end{equation}
Here, $AR_{i,j}$ denotes the probability that the $j^{th}$ draft token of the $i^{th}$ request is accepted. To unify speculative and autoregressive decoding, we define $AR_{i,0} = 1$, representing a bonus token that is always accepted.

\noindent\textbf{Modeling acceptance rate.} 
Acceptance in speculative decoding occurs when the draft model and the target model generate identical tokens at a given position, which implicitly implies high confidence from both models. Empirical observations (see Figure~\ref{fig:6}) confirm a strong correlation between confidence and acceptance. Similar to previous works~\cite{du2024glide,li2024eagle}, this finding suggests that we can model acceptance rates using confidence levels.

The probability that the $j^{th}$ draft token of the $i^{th}$ request is accepted depends on the successful acceptance of all previous tokens, and is approximated as:
\begin{equation}
\label{eq:ar_ij}
AR_{i,j}=\prod_{k=0}^{j}p_{i,k} \approx \prod_{k=0}^{j}c_{i,k}
\end{equation}
where $p_{i,k}$ is the conditional probability of the $k^{th}$ draft token of the $i^{th}$ request is accepted given that all previous draft tokens have been accepted, and $c_{i,k}$ is its confidence score. Notably, we define $p_{i,0} = c_{i,0} = 1$ to unify speculative and autoregressive decoding.

\begin{figure}[t]
\centering
\includegraphics[width=\linewidth]{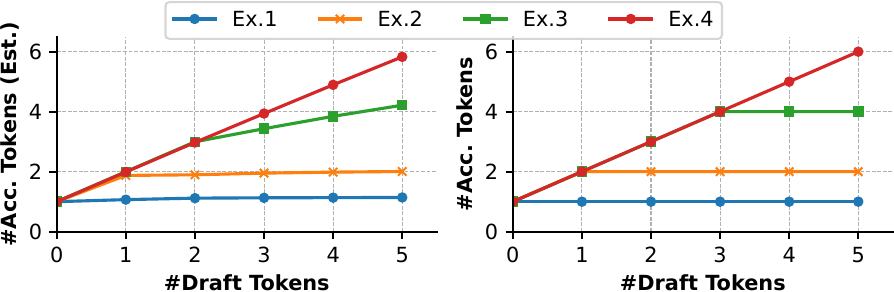}
\caption{The relationship between estimated and actual number of accepted token with speculative length.}
\label{fig:7}
\end{figure}
  
\noindent\textbf{The properties of the number of accepted tokens.} We analyze the mathematical properties of the average number of accepted tokens $\overline{NAT}$, based on the confidence-based acceptance model.

\begin{lemma}
\label{thm:avg_nat}
The average number of accepted tokens $\overline{NAT}$ has the following properties:
\begin{enumerate}
    \item (\emph{Initialization}) When $SL = 0$, $\overline{NAT} = 1$;
    \item (\emph{Marginal gain}) For $SL \geq 1$, the marginal gain is $\Delta \overline{NAT} \in [0,1]$;
    \item (\emph{Diminishing returns}) The marginal gain $\Delta \overline{NAT}$ is non-increasing as $SL$ increases.
\end{enumerate}
\end{lemma}

\begin{proof}
These results follow from the definition of ${NAT}$ in Eq.~\eqref{eq:nat} and the confidence-based approximation of $AR_{i,j}$ in Eq.~\eqref{eq:ar_ij}.

\begin{itemize}
    \item[(1)] When $SL = 0$, only $AR_{i,0} = 1$ contributes to the sum, hence $\overline{NAT} = 1$.
    
    \item[(2)] Increasing $SL$ by one adds an additional term $\Delta \overline{NAT} = \overline{AR_{SL}} = \frac{1}{BS} \sum_{i=1}^{BS} \prod_{k=0}^{SL} c_{i,k} \in [0,1]$, since each $c_{i,k} \in [0,1]$.
    
    \item[(3)] For each $i$ and $j$, we have $AR_{i,j+1} = AR_{i,j} \cdot c_{i,j+1} \leq AR_{i,j}$
    because $0 \leq c_{i,j+1} \leq 1$. Therefore, $AR_{i,j}$ is non-increasing in $j$, and so is its average $\overline{AR_j}$. Thus, $\Delta \overline{NAT} = \overline{AR_{SL}}$ is non-increasing.
\end{itemize}
\end{proof}
  
Fig. \ref{fig:7} confirms the mathematical properties of the accepted token numbers and shows that the difference between the estimated number of accepted tokens and the actual number of accepted tokens is minimal, proving the effectiveness of the confidence-based method for estimating the number of accepted tokens.

\subsection{The Relationship between Efficiency and Speculative Length}
\label{sec:relationship}

Let speculative length $SL$ be extended to a continuous variable $s \in \mathbb{R}_{\geq 0}$, and define the throughput function as:
\begin{equation}
f(s) = \frac{g(s)}{h(s)}
\end{equation}
where $g(s)$ is the average number of accepted tokens per request (i.e., $\overline{NAT}$), and $h(s)$ is the average speculative decoding time per request (i.e., $T_{sd}/BS$).

\noindent\textbf{Existence and location of optimal speculative length.} Due to the complexity of $f(s)$, we analyze its properties indirectly by leveraging the known properties of $g(s)$ and $h(s)$.

\begin{theorem}
\label{thm:existence}
$f(s)$ is quasi-concave on $\mathbb{R}_{\geq 0}$ and attains a unique global maximum $s^* \in \mathbb{R}_{\geq 0}$. Consequently, $f(s)$ is either monotonically decreasing, or strictly increasing on $[0, s^*)$ and strictly decreasing on $(s^*, \infty)$.
\end{theorem}

\begin{proof}
By Lemma~\ref{thm:avg_nat}, the numerator $g(s)$, which represents the average number of accepted tokens per request, is non-decreasing and concave, i.e., $g'(s) \geq 0$ and $g''(s) \leq 0$.  
By Lemma~\ref{lem:sd_time}, the denominator $h(s) = T_{sd}/BS$ is a quadratic function with positive coefficients, and hence strictly convex and increasing for $s \geq 0$.

The throughput function is defined as $f(s) = g(s)/h(s)$.  
Using standard quotient rule, its first derivative is:
\begin{align}
f'(s) = \frac{g'(s)h(s) - g(s)h'(s)}{\big(h(s)\big)^2}
\end{align}
and the second derivative is:
\begin{equation}
\begin{split}
f''(s) = & \frac{1}{\big(h(s)\big)^4} \Big( \big(g''(s)h(s) - g(s)h''(s)\big)\big(h(s)\big)^2 \\
& - 2h(s)h'(s)\big(g'(s)h(s) - g(s)h'(s)\big)\Big)
\end{split}
\end{equation}

Based on the derivative properties of $g(s)$ and $h(s)$, it follows that: (1) $f'(s)$ changes sign at most once from positive to negative, implying that $f(s)$ is either strictly decreasing or unimodal (increasing then decreasing) with a single peak; (2) $f(s)$ is quasi-concave on $s \in \mathbb{R}_{\geq 0}$ and attains a unique global maximum $s^* \in \mathbb{R}_{\geq 0}$..
\end{proof}

Specifically, the monotonicity of $f(s)$ is determined by the location of the optimal speculative length $s^*$.

\begin{corollary}
\label{cor:optimal_sl}
The optimal speculative length $s^*$ is positively correlated with the acceptance and the computational gap between the target and draft models, and negatively correlated with the batch size $BS$ and the average context length $\overline{N_c}$.
\end{corollary}

\begin{proof}
From the derivative expression in Theorem~\ref{thm:existence}, we know that $f'(0) > 0$ is equivalent to $g'(0)c - b > 0$, which determines whether the optimal length $s^* > 0$.

By substituting $c = h(0)$ and $b = h'(0)$ from Lemma~\ref{lem:sd_time}, we obtain:
\begin{align}
\begin{split}
g'(0)c - b = & 2\alpha_{v}\overline{N_{c}}g'(0) + 2\gamma_{v}g'(0) + \frac{2\delta_{v}}{BS}g'(0) - 2\alpha_{v}\overline{N_{c}} \\
& - \alpha_{v} - 2\gamma_{v} - 2\alpha_{d}\overline{N_{c}} + \alpha_{d} - 2\gamma_{d} - \frac{2\delta_{d}}{BS}
\end{split}
\end{align}

From this expression, we observe:
(1) A higher acceptance rate (which increases $g'(0)$) contributes positively, as it scales all positive terms;
(2) Reducing the batch size $BS$ increases the impact of both $\delta_v$ and $\delta_d$ terms due to their inverse dependence;
(3) A shorter average context length $\overline{N_c}$ decreases the cumulative weight of the $\alpha_v$ and $\alpha_d$ coefficients;
(4) Enlarging the computational gap (i.e., increasing $(\alpha_v, \gamma_v, \delta_v)$ and decreasing $(\alpha_d, \gamma_d, \delta_d)$) also favors a higher value of $g'(0)c - b$.

Hence, the optimal $s^*$ shifts to the right under these conditions.
\end{proof}

\noindent\textbf{Remark.}  
The structural property of $f(s)$ in Theorem~\ref{thm:existence} enables a practical tuning method. Since $f(s)$ is either monotonically decreasing or unimodal with a unique maximum, the optimal speculative length $s^*$ can be located by incrementally evaluating $f(s)$ at increasing speculative length and stopping at the first local maximum. This allows lightweight online tuning without exhaustive search.

\section{Adaptive Speculative Decoding}

Based on the relationship between efficiency and speculative length (\S\ref{sec:relationship}), we propose adaptive speculative decoding. 
To implement this framework, we developed three components. 
The first is the Adaptive Drafter (\S\ref{sec:adaptive_drafter}), which incorporates efficiency estimation into the drafting phase, achieving forward-looking step-level speculative length control. 
The second is the Confidence Prior Verifier (\S\ref{sec:confidence_prior_verifier}), which prioritizes the verification of tokens with high acceptance rates, achieving fine-grained request-level speculative length control. 
Lastly, the SLO-Aware Efficiency Estimator (\S\ref{sec:efficiency_estimator}) evaluates the efficiency of speculative decoding and achieves SLO awareness.

\subsection{Adaptive Drafter}
\label{sec:adaptive_drafter}

\begin{algorithm}[t]
    \renewcommand{\algorithmicrequire}{\textbf{Input:}}
    \renewcommand{\algorithmicensure}{\textbf{Output:}}
    \caption{Adaptive Drafter}
    \label{alg:1}
    \begin{algorithmic}[1]
        \REQUIRE Draft model $M_d$, Request set $reqs$, Confidence score of previous steps $prev\_conf$
        \ENSURE Draft token set $dtss$, Acceptance rate set $arss$
        \STATE $best\_thpt \gets -1$, $dtss \gets \{\}$, $arss \gets \{\}$
        \STATE $prev\_confs \gets \text{list}(prev\_conf) \times \text{lens}(reqs)$
        \STATE $arss.\text{append}(prev\_confs)$
        \WHILE{\textbf{True}}
            \STATE $thpt \gets \text{throughput\_estimation}(reqs,arss)$
            \IF{$thpt > best\_thpt$}
                \STATE $dts, confs \gets M_d(reqs, dtss)$
                \STATE $dtss.\text{append}(dts)$
                \STATE $arss.\text{pop}()$
                \IF{$arss == \{\}$}
                    \STATE $arss.\text{append}(confs)$
                \ELSE
                    \FOR{$i \gets 1$ to $\text{lens}(reqs)$}
                        \STATE $ars[i] \gets arss[\text{lens}(arss)-1][i] \times confs[i]$
                    \ENDFOR
                    \STATE $arss.\text{append}(ars)$
                \ENDIF
                \STATE $best\_thpt \gets \text{throughput\_estimation}(reqs,arss)$
                \STATE $arss.\text{append}(prev\_confs)$
            \ELSE
                \STATE $arss.\text{pop}()$
                \STATE \textbf{break}
            \ENDIF
        \ENDWHILE
    \end{algorithmic}
\end{algorithm}
  
Based on the relationship between efficiency and speculative length, we have developed a module named adaptive drafter. This module is capable of adaptively adjusting the number of executions of the draft model during the drafting phase, in response to changes in the generated information. To achieve this goal, we designed a speculative controller that estimates its acceptance rate by analyzing the confidence score of draft tokens and decides whether to continue executing the draft model based on changes in the efficiency of speculative decoding. If it is found that continuing execution would reduce the efficiency, the controller will halt further execution of the draft model. However, the challenge with this approach is that the speculative controller needs to estimate the efficiency after execution based on the generated confidence scores, leading to the necessity of an additional execution of the draft model to ensure optimal efficiency is achieved.

To address this challenge, we adopted a "predict-execute-correct" strategy. As demonstrated in Alg. \ref{alg:1}, we first use the average confidence scores from historical data to estimate the confidence scores that the next step of the draft model might produce, and based on this, predict the efficiency of speculative decoding. If the estimation indicates that continuing the execution of the draft model would improve the efficiency, the speculative controller executes the model, taking into consideration the newly generated draft tokens and their confidence scores, while also updating the collection of acceptance rates and the efficiency. If the estimation suggests that it is not appropriate to continue execution, the drafting phase is stopped. This approach enables the draft model to rapidly generate draft tokens that are more likely to be accepted.
\begin{algorithm}[t]
    \renewcommand{\algorithmicrequire}{\textbf{Input:}}
    \renewcommand{\algorithmicensure}{\textbf{Output:}}
    \caption{Confidence Prior Verifier}
    \label{alg:2}
    \begin{algorithmic}[1]
        \REQUIRE Target model $M_t$, Request set $reqs$, Draft token set $dtss$, Acceptance rate set $arss$
        \ENSURE Output token set $otss$
        \STATE $best\_thpt \gets \text{throughput\_estimation}(reqs, arss)$
        \WHILE{\textbf{True}}
            \STATE $min\_ar, min\_ar\_token \gets \text{find\_min\_ar}(dtss, arss)$
            \STATE $arss.\text{remove}(min\_ar)$
            \STATE $thpt \gets \text{throughput\_estimation}(reqs, arss)$
            \IF{$thpt > best\_thpt$}
                \STATE $best\_thpt \gets thpt$
                \STATE $dtss.\text{remove}(min\_ar\_token)$
            \ELSE
                \STATE \textbf{break}
            \ENDIF
        \ENDWHILE
        \STATE $otss \gets M_t(reqs, dtss)$
    \end{algorithmic}
\end{algorithm}

\subsection{Confidence Prior Verifier}
\label{sec:confidence_prior_verifier}

We developed a module named the confidence prior verifier, which introduces an innovative request-level speculative length control strategy by decoupling the drafting and verification phases of speculative decoding. Currently, speculative length control is typically implemented on step-level~\cite{liu2024optimizing}, which limits the ability to adjust speculative lengths for different requests within the same batch individually. The confidence prior verifier achieves request-level speculative length control indirectly by prioritizing the verification of draft tokens anticipated to have a higher acceptance rate, thereby validating different numbers of draft tokens for different requests. Considering that the verification phase may occupy up to 90\% of the decoding time, this differentiated verification strategy among requests efficiently realizes fine-grained speculative length control.

As illustrated in Alg. \ref{alg:2}, the confidence prior verifier employs a draft token elimination method to enhance the effectiveness of the draft token set entering the verification phase. This process starts with the collection of draft tokens and their estimated acceptance rates, evaluating whether a draft token should be eliminated in ascending order of acceptance rates. If the elimination operation can improve the efficiency of speculative decoding, the system executes the elimination and updates the collection of acceptance rates. This process is repeated until further elimination no longer enhances the efficiency. Ultimately, the Confidence Prior Verifier leverages the target model to validate the filtered draft tokens, achieving differentiated speculative decoding for different requests.

\subsection{SLO-Aware Efficiency Estimator}
\label{sec:efficiency_estimator}

\begin{algorithm}[t]
    \renewcommand{\algorithmicrequire}{\textbf{Input:}}
    \renewcommand{\algorithmicensure}{\textbf{Output:}}
    \caption{SLO-Aware Efficiency Estimator}
    \label{alg:3}
    \begin{algorithmic}[1]
        \REQUIRE Request set $reqs$, Acceptance rate set $arss$, SLO's TPOT constraint $tpot$
        \ENSURE Estimated goodput $gdpt$
        \STATE $t \gets \text{estimate\_time}(reqs)$
        \IF{$t > tpot$}
            \STATE $gdpt \gets -1$
        \ELSE
            \STATE $nat \gets \text{sum}(arss)$
            \STATE $gdpt \gets \frac{nat}{t}$
        \ENDIF
    \end{algorithmic}
\end{algorithm}
  
To support the adaptive drafter and the confidence prior verifier, we design a SLO-aware efficiency estimator (Alg. \ref{alg:3}). This estimator is capable of estimating the efficiency of speculative decoding based on the current workload and the speculative length.

Drawing from the model of accepted token numbers presented in \S\ref{sec:4.2}, the SLO-aware efficiency estimator utilizes the acceptance rate predicted by confidence to determine the number of tokens to be accepted at the current step. In conjunction with the speculative decoding time model outlined in \S\ref{sec:4.1}, it calculates the time required for speculative decoding based on the current workload. Through these two parameters, the estimator can estimate the speculative decoding throughput at the current step, thereby assisting \SysName in achieving the optimal efficiency of speculative decoding.

Furthermore, we guide \SysName in selecting the appropriate speculative length according to SLO requirements by setting constraints. Considering that in a batch of requests, some may only generate a single accepted token, making their TPOT equivalent to the time of speculative decoding. Therefore, to ensure that the time for each step of speculative decoding does not exceed the TPOT constraint specified by the SLO, strict control is necessary. If the time for speculative decoding exceeds the SLO limit, the throughput will be set to -1 to prevent the adaptive drafter from proceeding. In this manner, \SysName can adjust the speculative length based on SLO constraints to better meet the SLO.

\section{Implementation}

\SysName is implemented by extending the vLLM~\cite{10.1145/3600006.3613165} with approximately 2000 lines of Python code. vLLM is a well-known production-level serving system that efficiently manages memory through PagedAttention and integrates online inference techniques such as continuous batching. At the speculative decoding mode, the lookahead scheduler within the vLLM divides the memory into two equal parts, allocating one for the KV Cache of the draft model and the other for the KV Cache of the target model. When a new batch of requests comes in, the adaptive drafter performs the draft model autoregressively through an "predict-execute-correct" process, maintaining a collection of acceptance rates until generating a set of draft tokens with the optimal efficiency of speculative decoding at the step level. Following this, the confidence prior verifier eliminates some draft tokens with low acceptance rates that hinder the efficiency of speculative decoding, and then inputs the set of draft tokens with the optimal efficiency at the request level into the target model for verification. After generating the results of the speculative decoding for this step, these results are integrated into the input of the speculative decoding for the next step.

Furthermore, to model the speculative decoding time, we introduce an offline analyzer. After each change of the draft and target model pairs and computing platform, the offline analyzer analyzes the forward pass time of the draft and target models under different batch sizes and context lengths, using a linear regression model as \S\ref{sec:4.1} to derive performance coefficients. During subsequent operations, these performance coefficients, combined with the batch size and context length processed in each step of speculative decoding, are used to estimate the time for each step of speculative decoding.

\section{Evaluation}

\subsection{Experimental Setup}

\begin{table}[t]
\centering
\small
\renewcommand{\arraystretch}{1}
\caption{Model pairs and hardware.}
\begin{tabular}{ccc}
\hline
\textbf{Target Model (TP)} & \textbf{Draft Model (TP)} & \textbf{Hardware (VRAM)} \\ \hline
Vicuna-7B-v1.5 (1)         & Vicuna-68M (1)            & 1×L40 (48GB)             \\
                           &                           & 1×A100 (40GB)            \\ \hline
Vicuna-33B-v1.3 (4)        & Vicuna-160M (1)           & 4×A100 (160GB)           \\ \hline
Llama-3.3-70B (8)          & Llama-3.2-1B (1)          & 8×A100 (320GB)           \\ \hline
\end{tabular}
\label{tab:1}
\end{table}
  
\noindent\textbf{Model pairs and computing platforms.} We conducted tests on three model pairs, which were completed across four different computational platforms. Specifically, we selected three target models: Vicuna-7B-v1.5~\cite{zheng2023judging}, Vicuna-33b-v1.3, and Llama-3.3-70B-Instruct~\cite{dubey2024llama}, and matched them with Vicuna-68M~\cite{yang2024multi}, Vicuna-160M, and Llama-3.2-1B-Instruct as their respective draft models. Considering the memory resources, the model pair for the 7B target model operated on a single L40 or A100 (40GB) GPU, the model pair for the 33B target model operates on 4 $\times$ A100 (40GB) GPUs, while the model pair for the 70B target model runs on 8 $\times$ A100 (40GB) GPUs. In all these setups, the draft model operates with tensor parallelism set to 1, and only the target model is shared across multiple GPUs. We provide detailed specifications in Tab. \ref{tab:1}.
  
\noindent\textbf{Workloads.} To evaluate the performance of \SysName in real-world cloud inference systems, we selected three representative production traces from BurstGPT~\cite{burstgpt} and Mooncake~\cite{qin2025mooncake} to simulate request arrival patterns. 
To ensure the system load remains within normal operating ranges, we extracted 10-minute windows with moderate request rates for evaluation. As shown in Fig. \ref{fig:12}, these three traces simulate the following scenarios: (1) significant request rate variations (trace 1), (2) periodic request rate fluctuations (trace 2), and (3) rapid request rate oscillations (trace 3).

\begin{figure}[t]
\centering
\includegraphics[width=\linewidth]{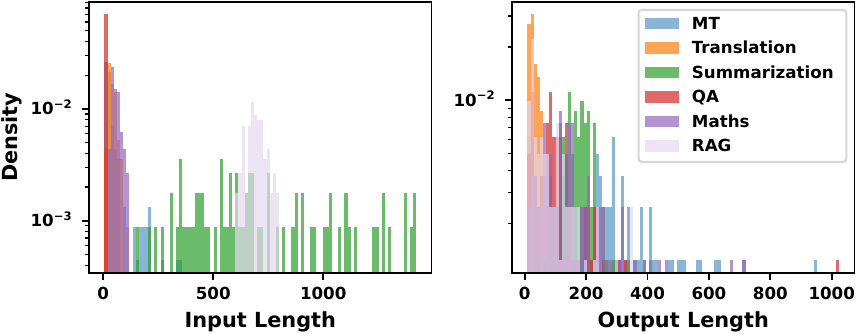}
\caption{Input and output length distributions corresponding to the Vicuna-7B-v1.5 model, showcasing the workload's key patterns.}
\label{fig:distribution}
\end{figure}

To accurately simulate the diversity of requests in online inference scenarios, we developed a mixed request set based on SpecBench~\cite{xia-etal-2024-unlocking} encompassing six distinct task categories: translation (WMT14 DE-EN), summarization (CNN/Daily Mail~\cite{nallapati-etal-2016-abstractive}), question answering (Natural Questions~\cite{10.1162/tacl_a_00276}), mathematical reasoning (GSM8K~\cite{cobbe2021training}), retrieval-augmented generation (DPR~\cite{karpukhin-etal-2020-dense}), and general dialogue (MTbench~\cite{zheng2023judging}). 
For each task category, we randomly sampled 80 instances from its corresponding dataset, totaling 480 request instances. 
Since the number of requests varies across traces, we sampled instances from this mixed set with equal probability to simulate real-world request patterns. 
Given that different target models employ distinct tokenizers and output modes, we present only the input and output length distributions corresponding to the Vicuna-7B-v1.5 model, as shown in Fig. \ref{fig:distribution}.

\begin{figure*}[t]
\centering
\includegraphics[width=\textwidth]{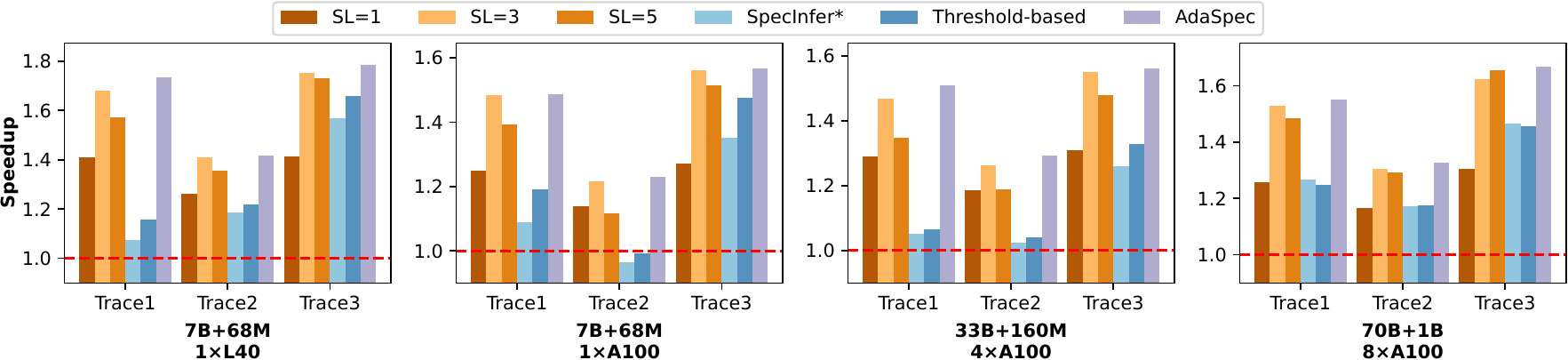}
\caption{Speedup in end-to-end latency across different traces, model pairs and computing platforms.}
\label{fig:8}
\end{figure*}

\begin{figure}[t]
\centering
\includegraphics[width=\linewidth]{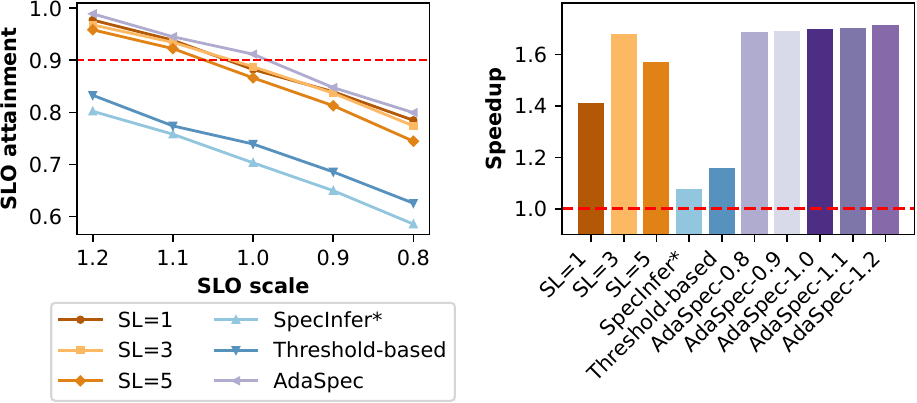}
\caption{SLO attainment and speedup in end-to-end latency across different traces, model pairs and computing platforms.}
\label{fig:9}
\end{figure}
  
\noindent\textbf{Baselines.} We compared \SysName with four baseline methods: (1) autoregressive decoding; (2) vanilla speculative decoding with preset speculative lengths of 1, 3, and 5; (3) threshold-based dynamic speculative decoding, a dynamic speculative decoding method that has been integrated into the transformer library~\cite{wolf-etal-2020-transformers}, representing the state-of-the-practice method; (4) SpecInfer~\cite{10.1145/3620666.3651335}, a state-of-the-art speculative decoding method integrated into inference systems. Due to the inefficiency of tree-based inference strategies (with a tree width of 3, the acceptance rate of draft tokens does not exceed 18\%~\cite{10.1145/3620666.3651335}), we set the tree width to 1 and name it SpecInfer*.

\noindent\textbf{Metrics.} 
To measure the latency reduction effect, we use the speedup of end-to-end latency relative to autoregressive decoding as our primary metric. 
Since our focus is on the benefits of speculative decoding in improving TPOT, we set the SLO for TPOT as the P90 TPOT value of autoregressive decoding, employing the SLO attainment rate to assess service reliability~\cite{zhong2024distserve,andes}. 
Furthermore, to evaluate the robustness of our approach under varying SLO constraints, we adjust the scale factor within a range of 0.8 to 1.2, simulating different levels of SLO strictness.

\subsection{Overall Performance of \SysName}
\label{sec:7.2}

\begin{figure}[t]
\centering
\includegraphics[width=\linewidth]{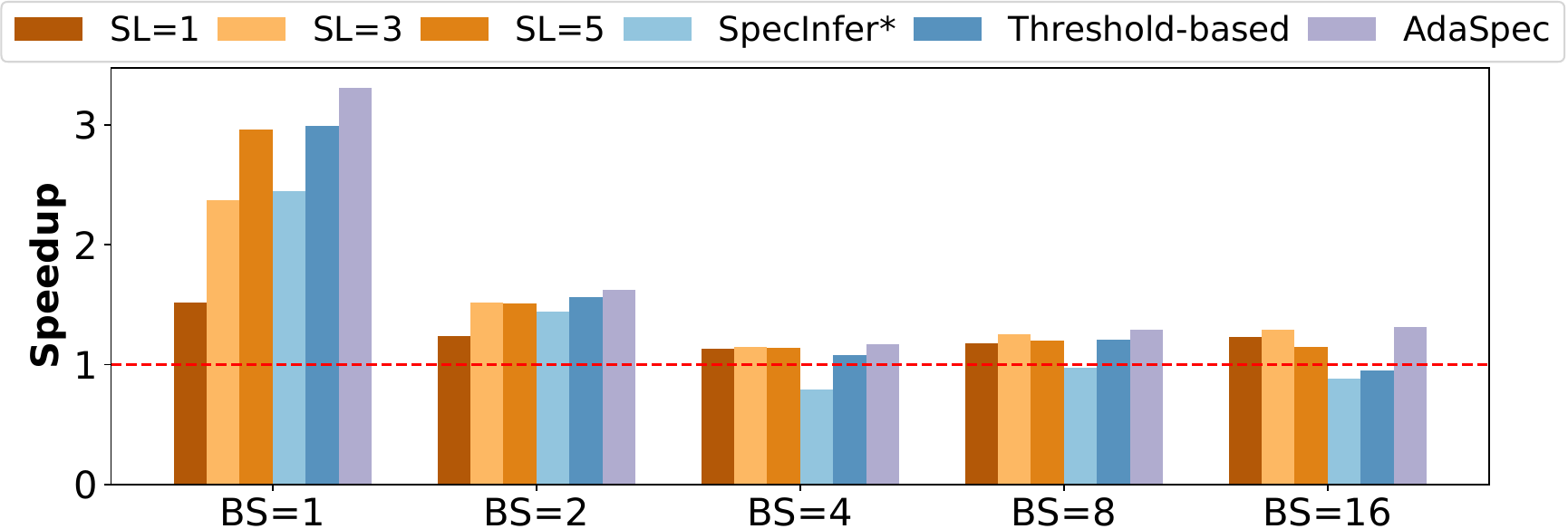}
\caption{Speedup in end-to-end latency across various batch sizes.}
\label{fig:10}
\end{figure}

As illustrated in Fig. \ref{fig:8}-\ref{fig:10}, we first evaluated the end-to-end latency speedup and SLO attainment of all methods under real-world production traces, and assessed the end-to-end latency speedup of all methods across various batch sizes.

\noindent\textbf{Consistent low inference latency.} 
As shown in Fig. \ref{fig:8}, \SysName demonstrates significant advantages across all 12 experimental configurations. 
Its core innovation lies in the fine-grained dynamic speculative length control mechanism that enables stable performance improvements across diverse complex scenarios. 
Compared to autoregressive decoding, \SysName achieves 1.23× to 1.79× speedup, proving the universal superiority of its speculative decoding framework. 
Against baseline methods with fixed speculative lengths, \SysName delivers an additional 7\%-18\% speedup through real-time speculative optimization, highlighting the critical value of its dynamic adjustment mechanism. 
The 16\%-66\% additional speedup over SpecInfer* reveals the inefficiency of verifying large batches of draft tokens in complex request patterns. 
Notably, \SysName achieves 9\% to 58\% greater acceleration than threshold-based dynamic methods, demonstrating that relying solely on confidence scores cannot accurately determine optimal speculative lengths, thus validating the advancement of \SysName's efficiency-optimized speculative decisions. 
By adapting to various request patterns and system configurations, \SysName consistently optimizes end-to-end latency, establishing itself as a reliable inference engine for real-world cloud production environments.

From a workload adaptability perspective, \SysName exhibits unique elastic capabilities: its 18\%-33\% speedup advantage in scenarios with drastic request rate fluctuations (trace 1) demonstrates exceptional adaptability to bursty loads; while maintaining stable 13\%-17\% speedup in rapidly changing environments (trace 3), confirming the robustness of its dynamic decision mechanism. 
Even in scenarios with pronounced periodic patterns (trace 2), \SysName breaks through the performance ceiling of baseline methods to achieve additional 10\%-14\% speedup, attributable to its fine-grained speculative length control strategy. 
Comparative analysis across these three scenarios shows \SysName effectively adapts to varying system loads, with performance benefits amplifying as load variability increases - a characteristic that makes \SysName particularly suitable for complex production environments.

In terms of system configurations, \SysName excels across hardware environments with varying arithmetic intensity: from 20\% performance improvement in 1×L40 configurations to consistent 15\% acceleration in 1×A100 setups. 
Its adaptive speculative strategy effectively overcomes the generalization limitations of traditional methods across hardware with different computational characteristics, making it an ideal choice for production deployment.

\begin{table*}[t]
\centering
\small
\caption{The variations in average speculative length, average number of accepted draft tokens, average draft token acceptance rate, and speedup across different traces for various dynamic methods.}
\begin{tabular}{cccccc}
\hline
\textbf{Trace}   & \textbf{Method} & \textbf{\# Draft Tokens (Avg.)} & \textbf{\# Accepted Draft Tokens (Avg.)} & \textbf{Acceptance Rate} & \textbf{SpeedUp} \\ \hline
\textbf{}        & Threshold-based & 5.97                            & 1.41                                     & 0.24                     & 1.16×            \\
\textbf{Trace 1} & \SysNameMinus        & 3.57                            & 1.22                                     & 0.34                     & 1.71×            \\
\textbf{}        & \SysName         & 3.42                            & 1.20                                     & \textbf{0.35}            & \textbf{1.73×}   \\ \hline
\textbf{}        & Threshold-based & 6.56                            & 1.42                                     & 0.22                     & 1.22×            \\
\textbf{Trace 2} & \SysNameMinus        & 3.13                            & 1.17                                     & 0.37                     & 1.40×            \\
\textbf{}        & \SysName         & 2.96                            & 1.14                                     & \textbf{0.39}            & \textbf{1.42×}   \\ \hline
\textbf{}        & Threshold-based & 5.26                            & 1.38                                     & 0.26                     & 1.66×            \\
\textbf{Trace 3} & \SysNameMinus        & 3.73                            & 1.26                                     & 0.34                     & 1.78×            \\
\textbf{}        & \SysName         & 3.63                            & 1.25                                     & \textbf{0.35}            & \textbf{1.79×}   \\ \hline
\end{tabular}
\label{tab:2}
\end{table*}

\noindent\textbf{High SLO attainment.} 
Fig. \ref{fig:9} presents the performance of various methods under a scenario with significant request rate fluctuations (trace 1) and varying SLO constraints. 
The experimental results demonstrate that \SysName exhibits significant advantages in achieving high SLO attainment. 
When the SLO constraint is set to the P90 TPOT value of autoregressive decoding (SLO scale=1), \SysName is the only method achieving an SLO satisfaction rate exceeding 90\%, with a speedup ratio over autoregressive decoding that outperforms other methods by 6\% to 66\%. 
Notably, under all SLO constraints, \SysName consistently achieves higher SLO satisfaction rates (up to 38\% improvement) while also delivering the most substantial speedup gains (up to 66\% improvement).

Furthermore, \SysName demonstrates unique adaptive capabilities, dynamically adjusting its speculative length based on different SLO constraints to optimize performance. 
For instance, under the most lenient SLO constraint (SLO scale=1.2), \SysName achieves an additional 2\% speedup compared to the strictest condition (SLO scale=0.8). 
This characteristic allows it to successfully address a critical challenge in speculative decoding---where pursuing higher speedup ratios typically leads to degraded SLO satisfaction rates. 
By employing SLO-aware adaptive speculative length control, \SysName not only maintains or surpasses the SLO satisfaction levels of autoregressive decoding but also achieves superior acceleration performance. 
As a result, it emerges as an ideal solution that balances both efficiency and SLO compliance.

\noindent \textbf{Batch adaptivity.} 
In dynamic production environments, fluctuations in batch size significantly impact system performance. 
As illustrated in Fig. \ref{fig:10}, \SysName demonstrates exceptional batch adaptability when handling arbitrary batch sizes with varying request compositions, achieving the highest speedup ratio among all baseline methods. 
This superior adaptability is reflected in the system's stable latency reduction and low-latency characteristics.

Compared to dynamic approaches, traditional fixed-length speculative decoding methods exhibit notable deficiencies in batch adaptability. 
These static methods typically face a dilemma: employing a conservative speculative length may maintain relatively stable latency reduction but fails to achieve optimal speedup, while adopting an aggressive speculative length may yield higher speedup at the cost of latency stability. 
Notably, even dynamic methods require an appropriate speculative length control mechanism to achieve optimal performance.
For instance, threshold-based dynamic methods can provide sporadic speedup improvements but struggle to maintain consistent latency reduction.

\SysName overcomes these limitations through its innovative dynamic adjustment mechanism. 
Its exceptional batch adaptability ensures efficient batch processing capability even under drastic batch size fluctuations---a critical feature for real-world production environments. 
This enables \SysName to deliver consistently stable and high-performance operation across diverse complex scenarios.

\subsection{Ablation Study}
\label{sec:7.3}

\noindent As shown in Tab. \ref{tab:2}, we conducted an ablation study of the performance of different dynamic methods on real-world production traces, exploring each method's speedup sources across three dimensions: average speculative length, average accepted draft tokens, and average draft token acceptance rate. Specifically, \SysNameMinus utilizes an adaptive drafter, without the confidence prior verifier.

\noindent\textbf{Benefits of adaptive drafter.} 
The adaptive drafter achieves significant performance improvements by dynamically controlling the speculative length, delivering an additional speedup ratio of 12\% to 55\%. 
Compared to traditional threshold-based methods, \SysNameMinus increases the average draft token acceptance rate by 8\% to 15\%, benefiting from its ability to promptly terminate the draft model's execution. 
The adaptive drafter can reduce the average speculative length (\# Draft Tokens (Avg.)) by 2.45 tokens while only decreasing the number of actually accepted draft tokens (\# Accepted Draft Tokens (Avg.)) by 0.19, indicating that \SysNameMinus's efficient acceleration primarily stems from its high conversion efficiency from draft tokens to accepted tokens. 
Clearly, compared to traditional threshold-based methods, the adaptive drafter employs confidence-based prediction model to precisely estimate the expected number of accepted tokens per batch. 
This enables finer-grained control over speculative length to improve token acceptance rates, and more efficient resource scheduling to enhance acceleration ratios.

\begin{figure}[t]
\centering
\includegraphics[width=\linewidth]{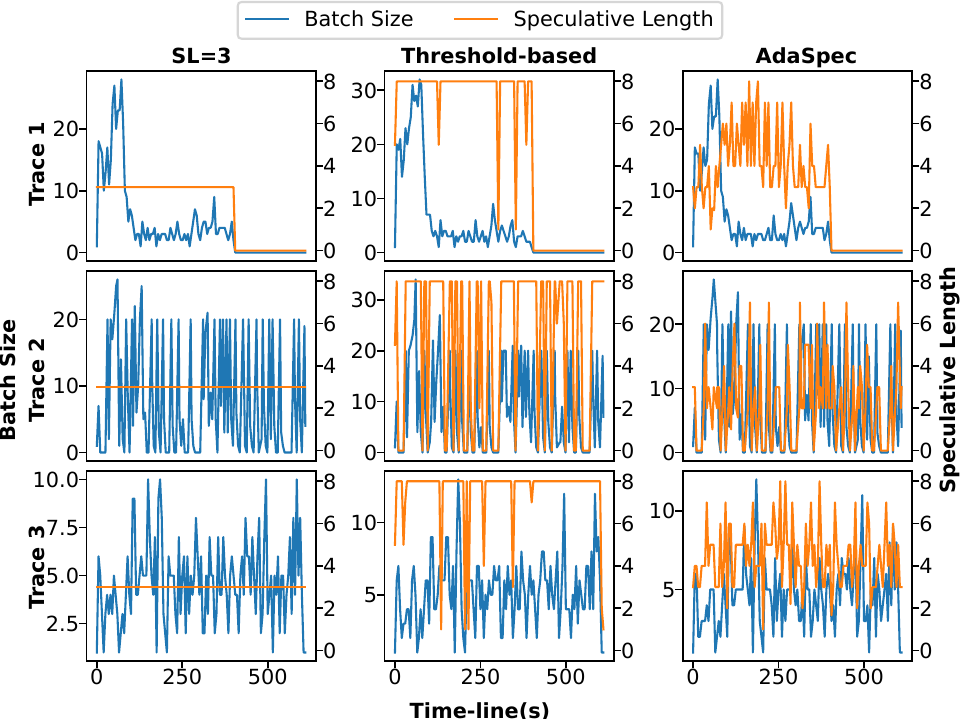}
\caption{Fluctuations in batch sizes and speculative lengths under different methods and traces.}
\label{fig:11}
\end{figure}

\noindent\textbf{Benefits of confidence prior verifier.} 
Request-level speculative length control further improves the speedup ratio. 
\SysName achieves the highest average draft token acceptance rate among all dynamic methods, which is the key reason for its superior acceleration performance. 
Compared to \SysNameMinus, \SysName slightly reduces the average number of accepted draft tokens (by only 0.015 on average) while also shortening the average speculative length (by 0.14 on average), ultimately increasing the average draft token acceptance rate by up to 2\%. 
This improvement demonstrates that the confidence prior verifier exhibits high selectivity when eliminating draft tokens, with only one out of every ten draft tokens likely being accepted, which confirms the effectiveness of the draft token elimination strategy.
As described above, the confidence prior verifier serves as a more fine-grained resource scheduling mechanism, further improving token acceptance rates and speedup ratios.

\subsection{Behavior Analysis}

\noindent As illustrated in Fig. \ref{fig:11}, we analyzed optimal vanilla speculative decoding (with a speculative length of 3), threshold-based dynamic speculative decoding, and \SysName across traces from three production environments. Within a 10-minute observation period, we monitored the batch size and speculative length every 5 seconds.

\noindent\textbf{Effective speculative length control.} 
In various production trace, \SysName exhibits a notable dynamic adjustment characteristic between its speculative length and batch size. 
Specifically, when the batch size increases with fluctuating request patterns, the speculative length decreases accordingly, and vice versa. 
This inverse regulation mechanism ensures efficient system operation under varying load conditions. 
Taking a scenario with significant request rate variations (trace 1) as an example, during high request rate phases, the system adapts to larger batch sizes (fluctuating between 10 and 20) by reducing the speculative length (varying between 2 and 4).
Conversely, during low request rate phases, it increases the speculative length (ranging from 4 to 7) to fully utilize smaller batch sizes (varying between 0 and 10), thereby optimizing resource efficiency based on real-time load conditions.

In contrast, threshold-based dynamic speculative decoding methods exhibit entirely different characteristics, with the speculative length consistently remaining near the preset maximum value of 8, almost unaffected by batch size variations. 
Although this approach simplifies the control strategy through straightforward confidence scores and threshold judgments, it results in a significantly higher average speculative length compared to \SysName, leading to resource wastage. 
Meanwhile, traditional fixed-length speculative decoding methods face even greater limitations. 
For instance, in most scenarios, the optimal preset speculative length (SL=3) performs similarly to \SysName’s speculative length (2-4) in the first half of trace 1. However, when the batch size drops sharply in the latter half of trace 1, the fixed length of 3 fails to fully exploit the potential of speculative decoding, unlike \SysName (with a speculative length of 4 to 7), ultimately increasing task completion latency.

Therefore, \SysName’s speculative length control mechanism demonstrates significant advantages in real-world production environments. 
Its ability to dynamically adjust computational resource allocation not only effectively reduces resource waste but also enhances the system’s capability to handle bursty requests.
By striking an optimal balance between performance and efficiency, \SysName provides an exemplary approach for practical cloud service systems.

\section{Related Work}

\noindent\textbf{Speculative decoding solutions.} The efficiency of speculative decoding relies on the draft model's speed and token acceptance rate, motivating research into fast draft models that closely approximate the target model.
There are two strategies: (1) Independent draft models, which involve introducing a high-efficiency independent draft model to generate draft tokens. This model could be a variant from the same series as the target model~\cite{pmlr-v202-leviathan23a,NEURIPS2023_6034a661}, a fine-tuned model~\cite{10.1145/3620666.3651335,NEURIPS2023_7b97adea,liu2024online,zhou2024distillspec}, or based on databases from other textual resources~\cite{he-etal-2024-rest}; 
(2) Self-draft models, which adjust the structure of the target model to enable it to rapidly generate multiple draft tokens~\cite{pmlr-v235-cai24b,lieagle,fu2024break,zhang-etal-2024-draft,li2024eagle}. 
For instance, the Multi-Token Prediction (MTP) adds multiple prediction heads to the target model, allowing it to generate consecutive draft tokens in parallel~\cite{liu2024deepseek}. 
In this paradigm, the speculative length can be interpreted as the number of active prediction heads.
\SysName offers universal adaptive control over the speculative length for both approaches, dynamically determining the optimal length for each request based on token-level confidence scores.
Contemporaneous with \SysName, AdaServe~\cite{li2025adaserve} also explores adaptive speculative decoding but differs: 
(1) AdaServe's primary objective is to satisfy the customized SLOs of each requests, whereas \SysName aims to balance generation latency with SLO attainment to ensure stable acceleration. 
(2) AdaServe targets tree-based speculative decoding, while \SysName is specifically optimized for sequential speculative decoding.

Additionally, speculative decoding has been adapted for specific use cases, such as processing ultra-long or medium-length sequences, by designing specialized draft models~\cite{sun2024triforce,chen2024magicdec}. 
This diversification of draft models has advanced the field, and \SysName provides fine-grained, adaptive control for any of them.

\noindent\textbf{Resource-efficient LLM serving.} Modern LLM serving systems use various orthogonal techniques to reduce latency and improve throughput.
For instance, Sarathi-Serve~\cite{sarathi-serve} maximizes throughput with chunked-prefills and stall-free scheduling. 
Prefill-Decoding (PD) Disaggregation~\cite{patel2024splitwise,qin2025mooncake,zhong2024distserve} assigns prefill and decoding phases to separate GPUs to meet their respective SLOs. 
For multi-tenant scenarios, Llumnix~\cite{298685} optimizes cross-node resources through runtime rescheduling. 
\SysName further enhances efficiency by introducing fine-grained speculative length control, enabling speculative decoding as a stable and high-performance decoding solution. 
These techniques operate orthogonally, collectively optimizing resource utilization in LLM serving systems.


\section{Conclusion}

Speculative decoding aims to reduce inference latency by increasing computational overhead, yet this trade-off depends on the variations in request patterns, system configurations, and SLOs. We explore how these variables influence speculative decoding and have developed \SysName, an adaptive speculative decoding framework that addresses these variations through request-level speculative length control. Experimental results demonstrate that \SysName can effectively and consistently reduce inference latency while maintaining lower computational costs.

\begin{acks}
We thank the anonymous reviewers for their insightful comments that helped improve this work. We also thank Zhaoyu Fan, Bowen Han, Zehua He, and Chengyue Wu for their helpful comments. This work was supported in part by the Science and Technology Commission of Shanghai Municipality (24DP1500704 and 24YL1901100), the Fundamental Research Funds for the Central Universities (22120230311), and the Guangdong Provincial Key Laboratory of Mathematical Foundations for Artificial Intelligence (2023B1212010001).
\end{acks}

\balance
\bibliographystyle{ACM-Reference-Format}
\bibliography{sample-base}


\end{document}